\documentclass{article}
\usepackage{spconf,amsmath,graphicx}

\usepackage{subcaption}
\usepackage{algorithm}
\usepackage{algpseudocode}
\usepackage{amsfonts} 
\usepackage{hyperref}

\newcommand\blfootnote[1]{%
  \begingroup
  \renewcommand\thefootnote{}\footnote{#1}%
  \addtocounter{footnote}{-1}%
  \endgroup
}

\title{The rise of the lottery heroes: why zero-shot pruning is hard}
\name{Enzo Tartaglione}
\address{LTCI, T\'el\'ecom Paris, Institut Polytechnique de Paris}
\begin{document}
%
\maketitle
\begin{abstract}
Recent advances in deep learning optimization showed that just a subset of parameters are really necessary to successfully train a model. Potentially, such a discovery has broad impact from the theory to application; however, it is known that finding these trainable sub-network is a typically costly process. This inhibits practical applications: can the learned sub-graph structures in deep learning models be found at training time?\\
In this work we explore such a possibility, observing and motivating why common approaches typically fail in the extreme scenarios of interest, and proposing an approach which potentially enables training with reduced computational effort. The experiments on either challenging architectures and datasets suggest the algorithmic accessibility over such a computational gain, and in particular a trade-off between accuracy achieved and training complexity deployed emerges.
\end{abstract}
\begin{keywords}
The lottery ticket hypothesis, pruning, computational complexity, deep learning\blfootnote{Accepted for publication at the IEEE International Conference on Image Processing (IEEE ICIP 2022).\\~\\ © 2022 IEEE. Personal use of this material is permitted. Permission from IEEE must be obtained for all other uses, in any current or future media, including reprinting/republishing this material for advertising or promotional purposes, creating new collective works, for resale or redistribution to servers or lists, or reuse of any copyrighted component of this work in other works.}
\end{keywords}

\section{The elephant in the room}
\label{sec:introduction}

Artificial neural networks (ANNs) are nowadays one of the most studied algorithms used to solve a huge variety of tasks. Their success comes from their ability to learn from examples, not requiring any specific expertise and using very general learning strategies. However, deep models share a common obstacle: the large number of parameters, which allows their successful training~\cite{ba2014deep,denton2014exploiting}, determines high training costs in terms of computation. For example, a ResNet-18 trained on ILSVRC'12 with a standard learning policy~\cite{resnetinet}, requires operations in the orders of hundreds of PFLOPs for back-propagation, or even efficient architectures like MobileNet-v3~\cite{howard2019searching} on smaller datasets like CIFAR-10 with an efficient learning policy~\cite{mobnetcifar10}, require order of hundreds of TFLOPs for back-propagation! Despite an increasingly broad availability of powerful hardware to deploy training, energetic end efficiency issues still need to be addressed.
\begin{figure}
    \centering
    \includegraphics[width=0.6\columnwidth, trim={0 130 0 0},clip]{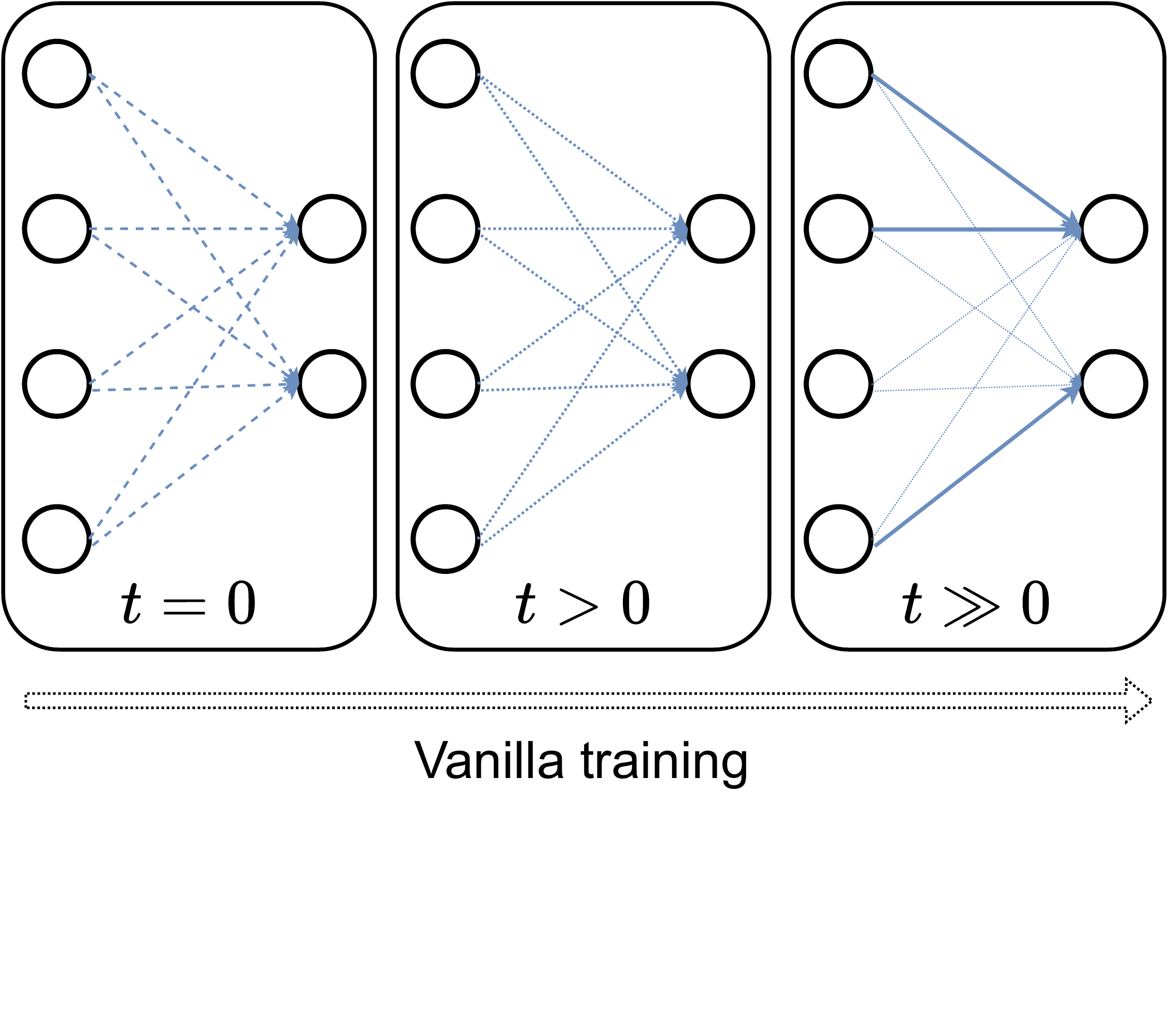}
    \caption{A subset of parameters, sufficient to reach good generalization, is typically determined in an iterative fashion. Can they be determined earlier, during a normal vanilla training?}
    \label{fig:idea}
\end{figure}

\noindent Some approaches have been proposed in order to reduce the computational complexity for deep neural networks. The act of removing parameters (or entire units) from a deep neural network is named \emph{pruning}. Despite the first works have been proposed many decades ago~\cite{lecun1990optimal}, pruning became popular just a few years ago, targeting the reduction of the model's size at deployment time and making inference more efficient~\cite{han2015learning,molchanov2017variational,bragagnolo2021role,luo2017thinet,tartaglione2021serene}.\\
A recent work, the \emph{lottery ticket hypothesis}~\cite{frankle2018lottery}, suggests that the fate of a parameter, namely whether it is useful for training (winner at the lottery of initialization) or if it can be removed from the architecture, is decided already at the initialization step. Frankle and Carbin propose experiments showing that, with an a-posteriori knowledge of the training over the full model, it is possible to identify these parameters, and that it is possible to successfully perform a full training just with them, matching the performance of the full model. However, in order to identify these winners, a costly iterative pruning strategy is deployed, meaning that the complexity of finding the lottery winners is larger than training the full model. Is it possible to deploy a zero-shot strategy, where we identify the lottery winners before, or during, the training of the model itself, to get a real computational advantage?
\begin{figure*}
\centering
\begin{subfigure}[b]{0.233\textwidth}
    \includegraphics[width=\columnwidth, trim={5 5 70 5},clip]{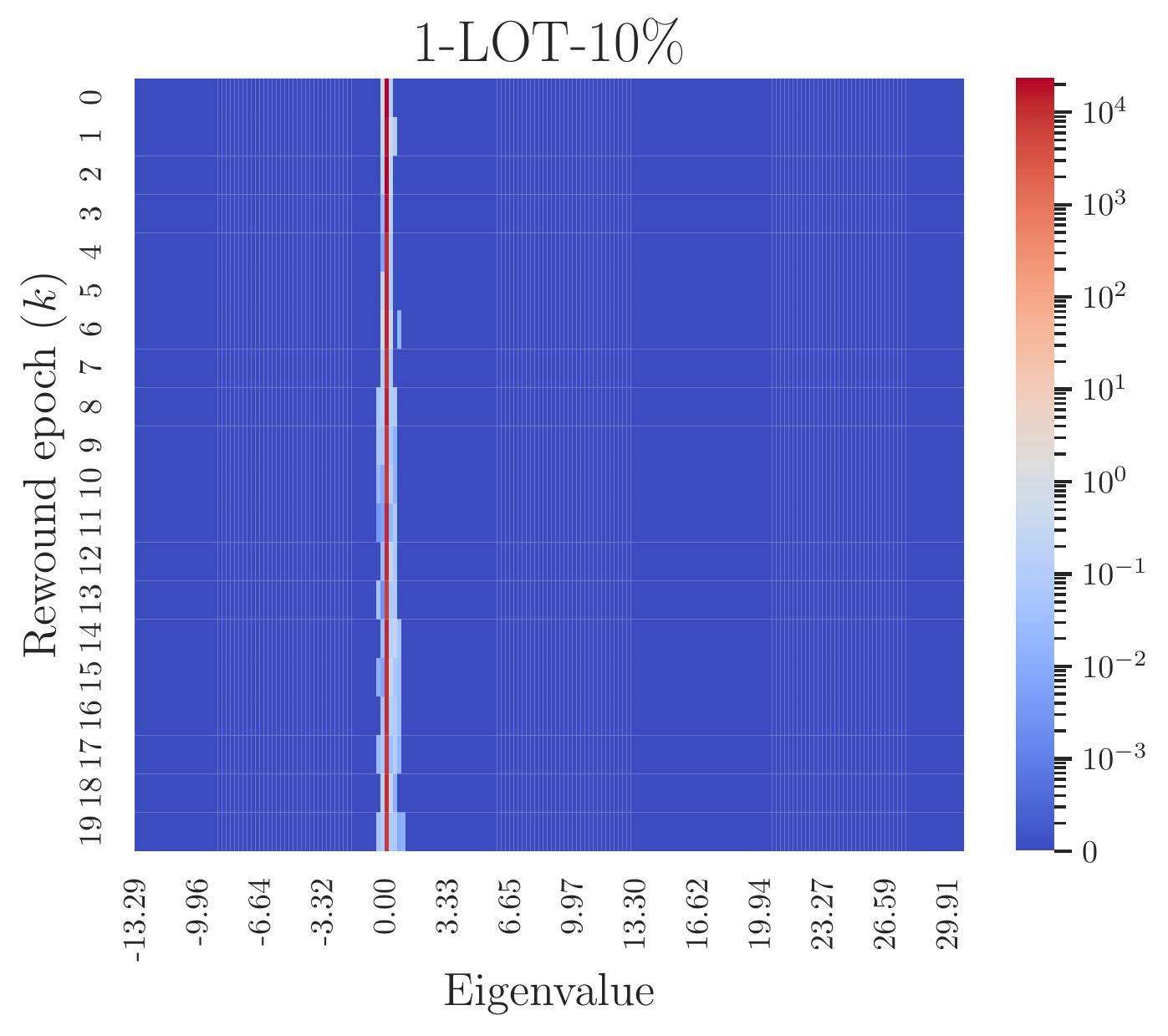}
    \caption{~}
    \label{fig:hess10}
\end{subfigure}
\begin{subfigure}[b]{0.233\textwidth}
    \includegraphics[width=\columnwidth, trim={5 5 70 5},clip]{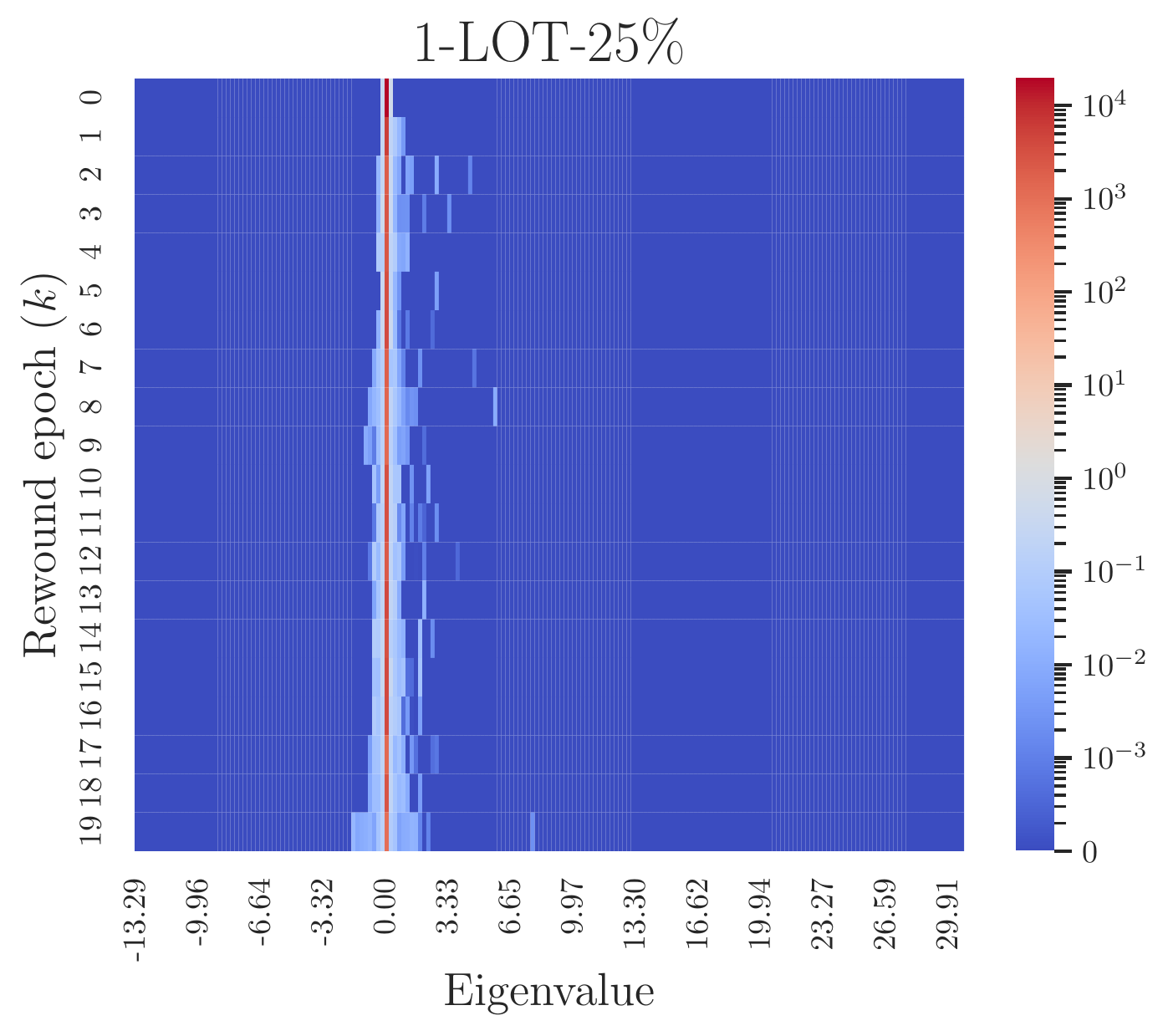}
    \caption{~}
    \label{fig:hess25}
\end{subfigure}
\begin{subfigure}[b]{0.233\textwidth}
    \includegraphics[width=\columnwidth, trim={5 5 70 5},clip]{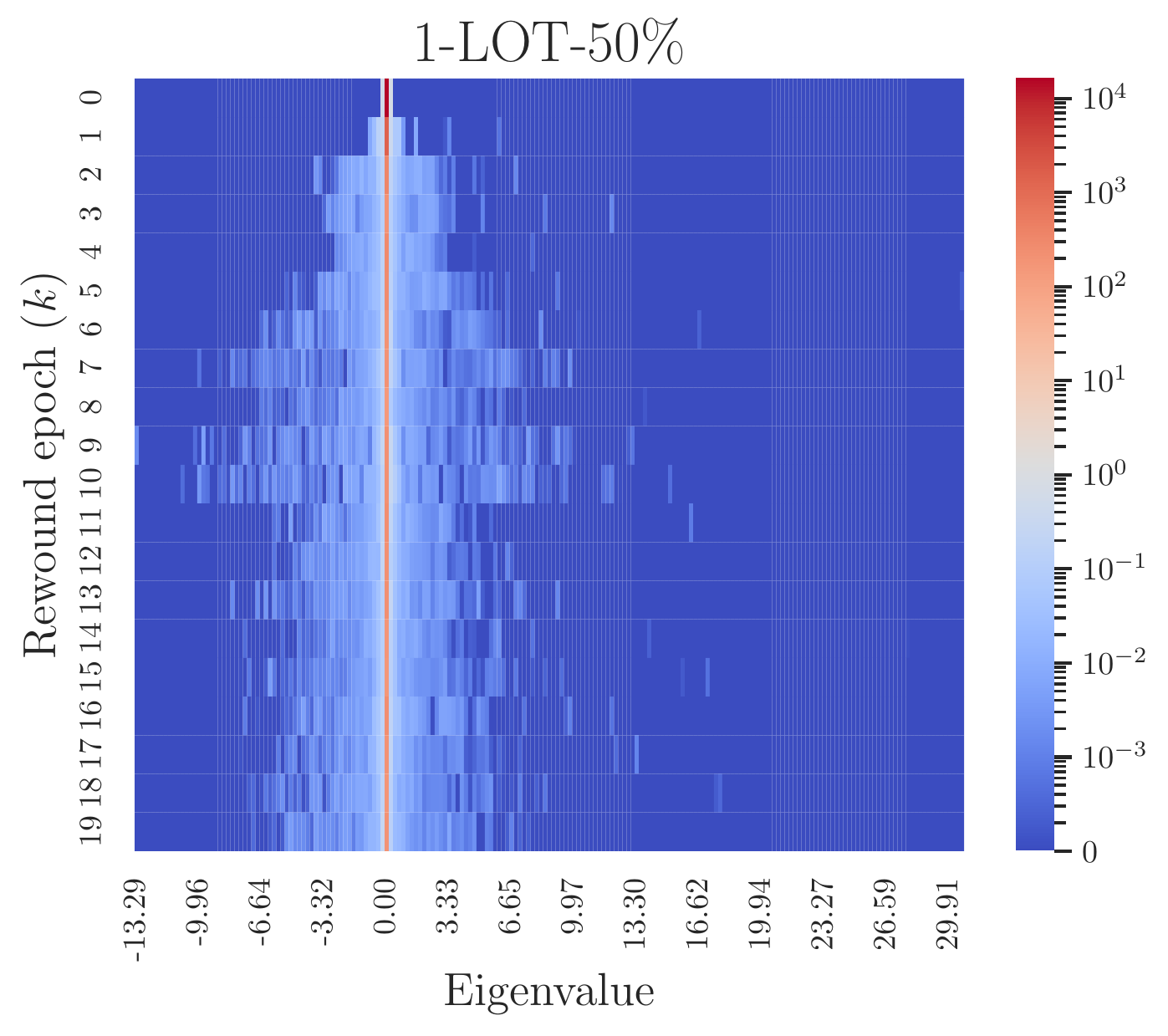}
    \caption{~}
    \label{fig:hess50}
\end{subfigure}
\begin{subfigure}[b]{0.278\textwidth}
    \includegraphics[width=\columnwidth, trim={5 5 5 5},clip]{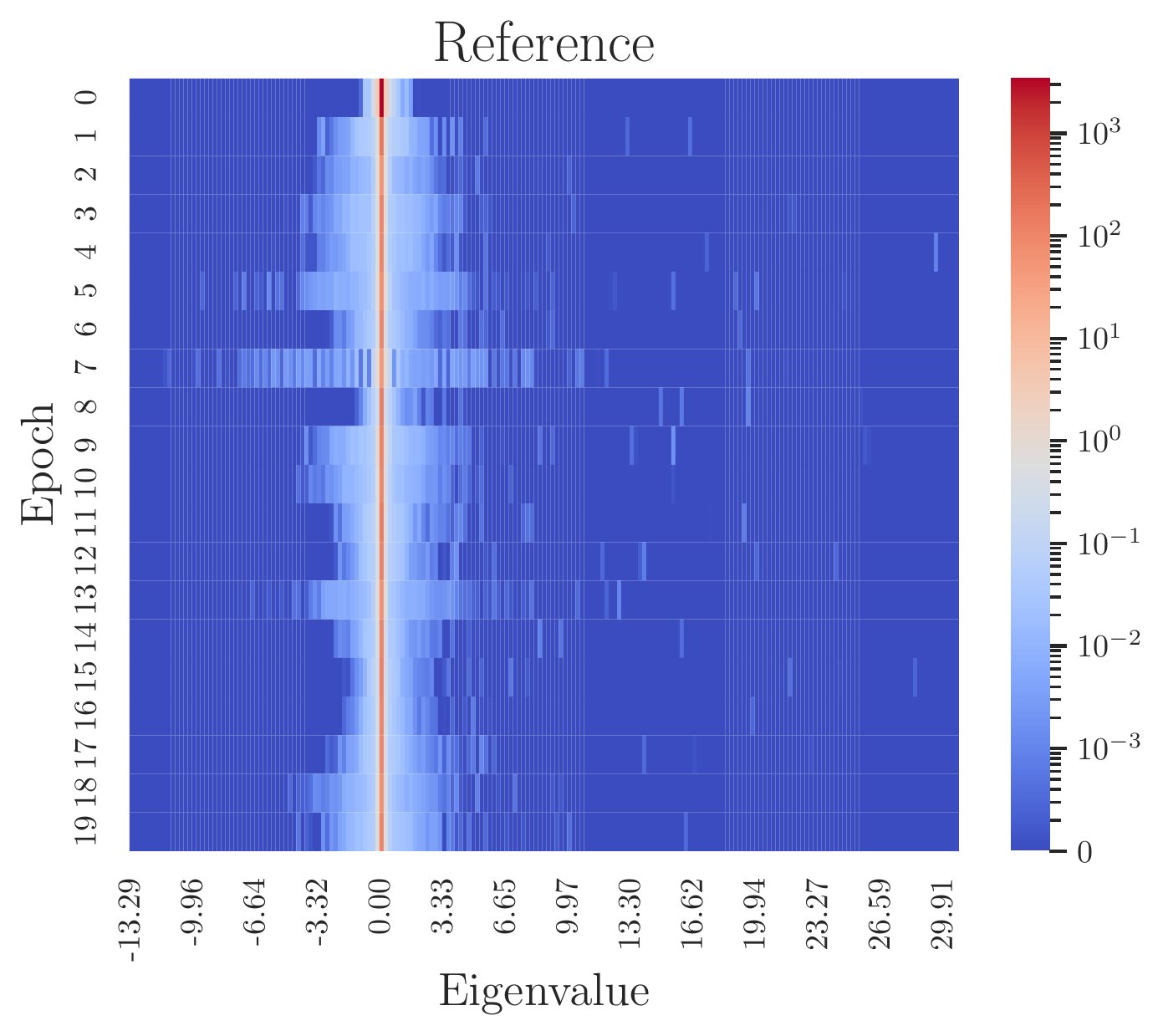}
    \caption{~}
    \label{fig:hessref}
\end{subfigure}
\caption{Example of distribution of the eigenvalues of the Hessian matrix calculated on the CIFAR-10 training set for ResNet-32 along different rewound epochs ($k$) retaining the 10\% of the parameters (a), the 25\% (b) the 50\% (c) and all the parameters (d). Here $I=1$. The represented scenario is qualitatively matched for different initialization of the model.}
\label{fig:hessianspecanalysis}
\end{figure*}

\noindent In this work we ground the lottery ticket hypothesis, motivating why the originally proposed strategy, despite showing the existence of the lottery tickets, is computationally sub-optimal. We leverage over experiments on CIFAR-10 and ILSVRC'12, qualitatively and quantitatively, analyzing the loss landscape evolution and proposing a strategy which opens the road to the design of optimization strategies which can effectively save computational power at training time. The lottery tickets are not evident in the first epochs, but they rise when the model's parameters have reached a specific subspace, and that iterative pruning strategies, which are necessary for traditional lottery ticket approaches, are not necessary to identify the lottery winners (Fig.~\ref{fig:idea}). We observe the feasibility of having a pruning strategy on-going at training time, and that, in very high compression regimes, the performance is mainly bound by the computational complexity budget we are willing to deploy. 
\section{The lottery of the initialization}
\label{sec:lottery}


\textbf{The lottery ticket hypothesis.} It is a known fact that deep neural networks are typically over-parametrized and, after training, a part of the parameters can be removed without harming the performance, or even slightly improving the performance in small pruning regimes~\cite{han2015learning,tartaglione2020pruning}. However, an interesting question rises: is it possible to train a sub-network, still achieving the same performance as training the full model? In their famous paper, Frankle and Carbin provide a method to identify, from a the set of initialized parameters, a subset $\mathcal{W}$ of the parameters which are sufficient, when trained in isolation, to achieve the same performance of the full model~\cite{frankle2018lottery}. From a practical perspective, this means that all the parameters not in $\mathcal{W}$ (hence in $\overline{\mathcal{W}}$) are pruned from the model, or more concretely, their value is locked to zero. Let us say that a trained model has $N$ parameters: we define a mask $\mathcal{M}\in \{0;1\}^N$ of the same dimensionality of the original model's parameters $W\in \mathbb{R}^N$ and we say that a parameter $w_i\in \mathcal{W} \Leftrightarrow m_i=1| m_i\in \mathcal{M}$.\footnote{we index the parameters of the models along a unique vector for simplicity.}
\begin{algorithm}
\caption{Lottery winners in $I$ iterations with $R\%$ remaining parameters at every iteration (I-LOT-R).}
\label{alg:lottery}
\begin{algorithmic}[1]
\Procedure{I-LOT-R($W^{0}$, $R$, $I$)}{}
\State $i \gets 0$
\State $\mathcal{M} \gets 1$\Comment{unit vector}
\While {$i < I$}
    \State $W_{LOT}^{0} \gets W^{0} \cdot \mathcal{M}$\label{line:rewind}
    \State $W_{LOT}^{f} \gets $ \Call{Train($W_{LOT}^{0}$, $\mathcal{M}$)}{}\Comment{\eqref{eq:updatelottery}}\label{line:train}
	\State $\mathcal{M} \gets $\Call{Magnitude prune($W_{LOT}^{f}$, $R$, $\mathcal{M}$)}{}\label{line:IMP}
    \State $i \gets i+1$
\EndWhile
\State \Return $\mathcal{M}$
\EndProcedure
\end{algorithmic}
\end{algorithm}
Alg.~\ref{alg:lottery} reports the algorithm to find the lottery winners, which involves an iterative magnitude pruning strategy (IMP, line~\ref{line:IMP}): after every training round (line~\ref{line:train}) the lowest $(100-R)\% \in \mathcal{W}$ having the smallest magnitude will be removed from $\mathcal{W}$. The parameters in $\mathcal{W}$ will then be \emph{rewound} to their original values (line~\ref{line:rewind}) and a new training, just updating $\mathcal{W}$, will be performed:
\begin{equation}
    \label{eq:updatelottery}
    w_{i}^{t+1} = \left\{
    \begin{array}{ll}
    w_{i}^{t} - u_{i}^{t} & if~w_i \in \mathcal{W}\\
    0   & if~w_i\in \overline{\mathcal{W}},
    \end{array}
    \right .
\end{equation}
where $u_{i}$ is some generic update term. In principle, the parameters in $\overline{\mathcal{W}}$ are not in the model, and for instance they should not be included in the computation anymore; however, we still need to encode that are missing, producing an overhead, as they are removed in an unstructured way~\cite{bragagnolo2021role}.\footnote{unless entire structures are not entirely removed from the model, but this is not the general case.}
\begin{algorithm}
\caption{Lottery winners with $k$ epochs warm-up.}
\label{alg:warmup}
\begin{algorithmic}[1]
\Procedure{I-LOT-R with warm-up($W^{0}$, $R$, $I$, $k$)}{}
\State $e \gets 0$
\State $W^e \gets W^{0}$
\While {$e < k$}
    \State $W^e \gets $ \Call{Train one epoch($W^e$)}{}\Comment{Here all the weights are trained, for one epoch only}
    \State $e \gets e+1$
\EndWhile
\State $W^k \gets W^e$
\State $\mathcal{M} \gets$ \Call{I-LOT-R($W^k$, $R$, $I$)}{}\label{line:calling_method}
\State \Return $\mathcal{M}$
\EndProcedure
\end{algorithmic}
\end{algorithm}
\begin{algorithm}
\caption{Rise of the lottery heroes with $R\%$ remaining parameters (RISE-R).}
\label{alg:rise}
\begin{algorithmic}[1]
\Procedure{RISE-R($W^{k}$, $R$)}{}
\State $W^f \gets $ \Call{Train($W^k$)}{}
\State $\mathcal{M} \gets $\Call{Magnitude prune($W^f$, $R$)}{}
\State $W_{RISE}^f \gets $ \Call{Train($W^k$, $\mathcal{M}$)}{}\Comment{\eqref{eq:updaterise}}
\State \Return $\mathcal{M}$
\EndProcedure
\end{algorithmic}
\end{algorithm}

\noindent\textbf{Limits.} Despite achieving the purpose of showing that winning tickets exist, there is a major, significant drawback of the approach in Alg.~\ref{alg:lottery}: the complexity of the overall strategy, namely the number of rewinds $I$ to converge to the target minimal subset $\mathcal{W}$, which depends on the amount of remaining parameters $R$. Such a value can not be set to very high values, as the approach fails. In order to improve this aspect, more works have tried to address possible solutions. In particular, \cite{frankle2020linear} shows that there is a region, at the very early stages of learning, where the lottery tickets identified with iterative pruning are not stable (if they are found, for different seeds they are essentially different). The novelty here introduced is an inspection over the epoch (or mini-batch iteration) where to rewind: simply, we pass to Alg.~\ref{alg:lottery} the parameters of a model already trained for the first $k$ epochs (Alg.~\ref{alg:warmup}). This is endorsed also by other works, like \cite{frankle2020early,morcos2019one,malach2020proving,girish2021lottery}, while other works reduce the overall complexity of the iterative training by drawing early-bird tickets~\cite{you2019drawing} (meaning that they learn the lottery tickets when the model have not yet reached full convergence) or even reducing the training data~\cite{zhang2021efficient}.\\
\textbf{Preliminary experiment and analysis.} The golden mine in this context would be to address a strategy for zero-shot lottery drafting, meaning that the lottery tickets are identified before the training itself. In order to assess its feasibility, let us define a companion model (ResNet-32) trained on CIFAR-10 for 180 epochs, using SGD optimization with initial learning rate 0.1 and decayed by a factor 0.1 at milestones 80 and 120, with momentum 0.9, batch size 100 and weight decay $5\cdot 10^{-5}$, as in~\cite{ResNetcifar10}. Fig.~\ref{fig:hessianspecanalysis} reports the distribution of the eigenvalues of the Hessian computed for the first 19 epochs on $W_{LOT}^{k} = W^{k} \cdot \mathcal{M}$ (Alg.~\ref{alg:warmup}) with $I=1$, evaluated on the full training set. We are interested in this specific one-shot scenario as we explore the \emph{possibility} of removing trained parameters \emph{during training} towards computational complexity saving, and in the one-shot scenario we remove them from the original, vanilla training trajectory. For this experiment, the \texttt{pyhessian} library has been used~\cite{yao2020pyhessian}, along with a NVIDIA~A40 GPU. We observe that, compared to the reference (namely, the distribution of the eigenvalues evaluated on the full model - Fig.~\ref{fig:hessref}) when $R$ is low ($R=10\%$ - Fig.~\ref{fig:hess10} - or $R=25\%$ - Fig.~\ref{fig:hess25}), the distribution changes significantly. In particular, a peak to values close to zero is observed: locally, the loss landscape is flat. Contrarily, for a higher $R$ regime (Fig.~\ref{fig:hess50}) the distribution is richer and similar to the reference (Fig.~\ref{fig:hessref}). When the loss landscape becomes flatter, the optimization problem itself is harder. We observe indeed that, with respect to a baseline performance of 92.92\% on the test set, with $R=10\%$, despite rewinding up to $k=20$, the achieved performance is never above 60\%. Why does this happen? In the next section we tackle this problem motivating why it is hard to evaluate the winning tickets when $I=1$ (or simply, in a one-shot fashion).

\section{Winning tickets in hindsight: the rise of the lottery heroes}
\label{sec:riseofshieldhero}
\begin{figure}
    \centering
    \includegraphics[width=0.63\columnwidth, trim={85 50 80 100},clip]{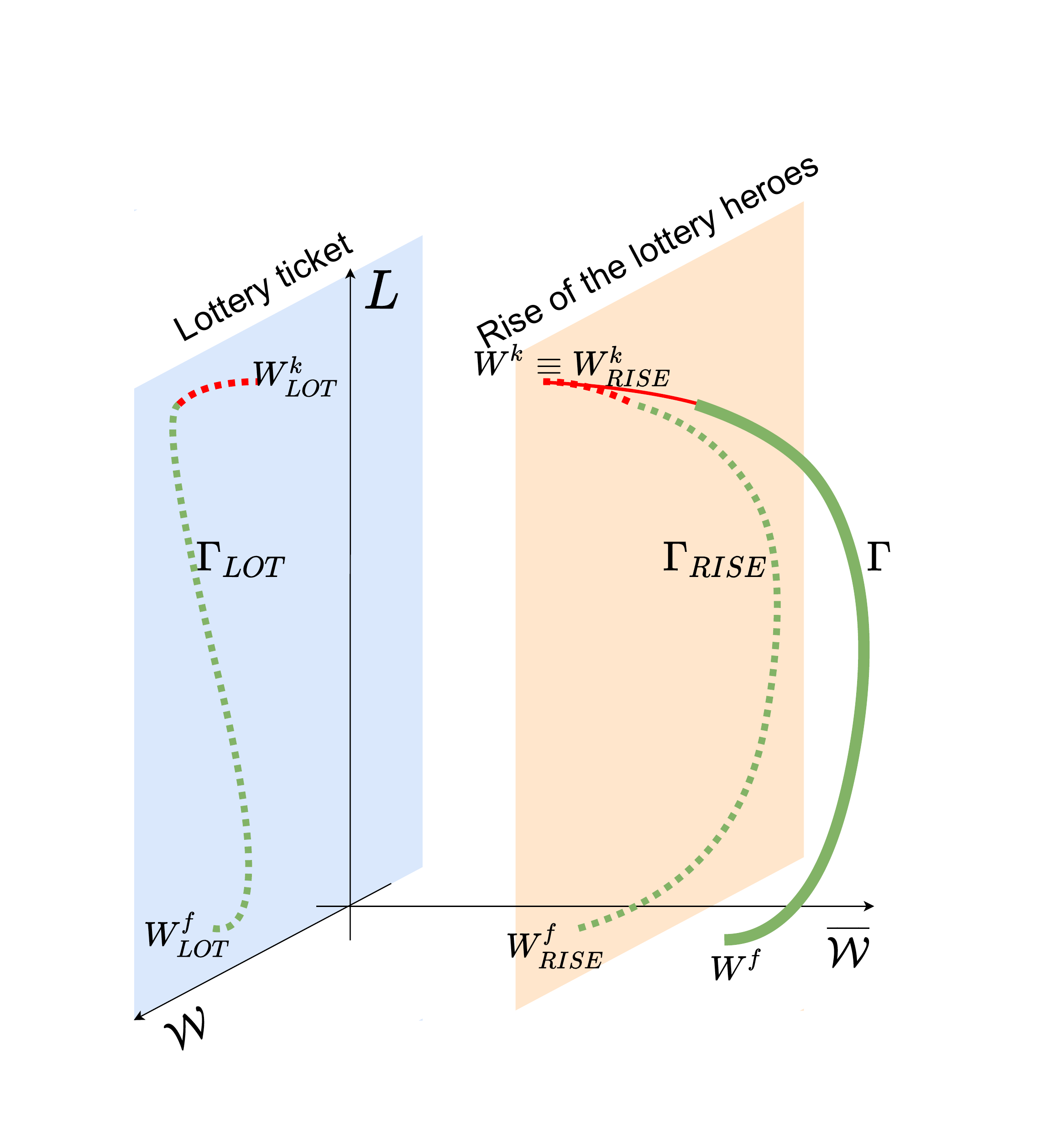}
    \caption{LOT projects the parameters in the subspace $\mathcal{W}$: for low $R$ the learning trajectory $\Gamma_{LOT}$ is very steep making the optimization problem hard, compared to the trajectory $\Gamma$ of the full model. RISE, on the contrary, does not project the parameters, but constrains the optimization problem to the parameters identified by $\mathcal{M}$.}
    \label{fig:intuition}
\end{figure}
\begin{figure}[t!]
\centering
\begin{subfigure}{\columnwidth}
    \includegraphics[width=0.99\columnwidth, trim={5 5 8 5},clip]{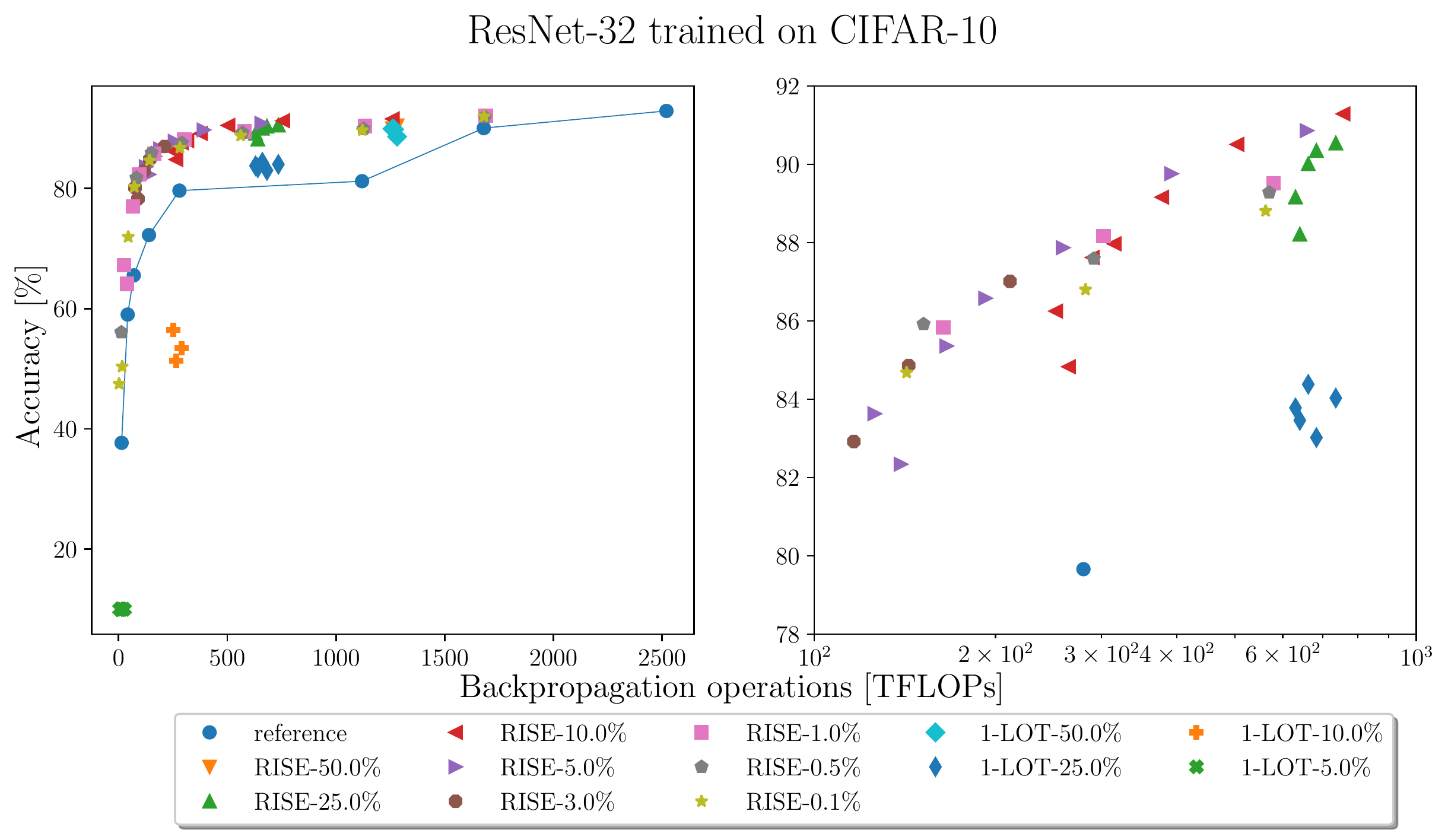}
\end{subfigure}
~\\~\\
\begin{subfigure}{\columnwidth}
    \includegraphics[width=0.99\columnwidth, trim={7 5 5 5},clip]{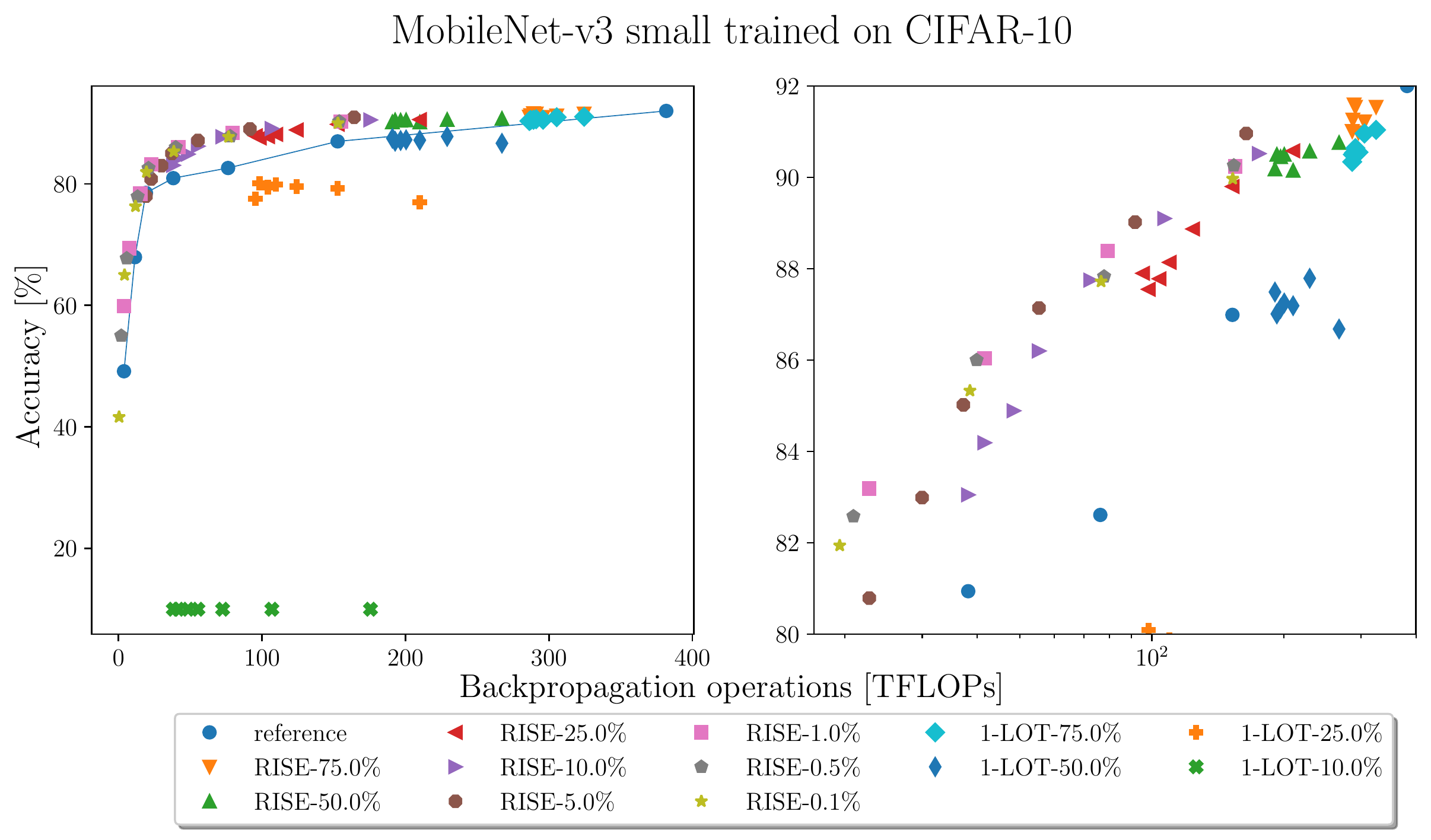}
\end{subfigure}
~\\~\\
\begin{subfigure}{\columnwidth}
    \includegraphics[width=0.99\columnwidth, trim={5 5 5 5},clip]{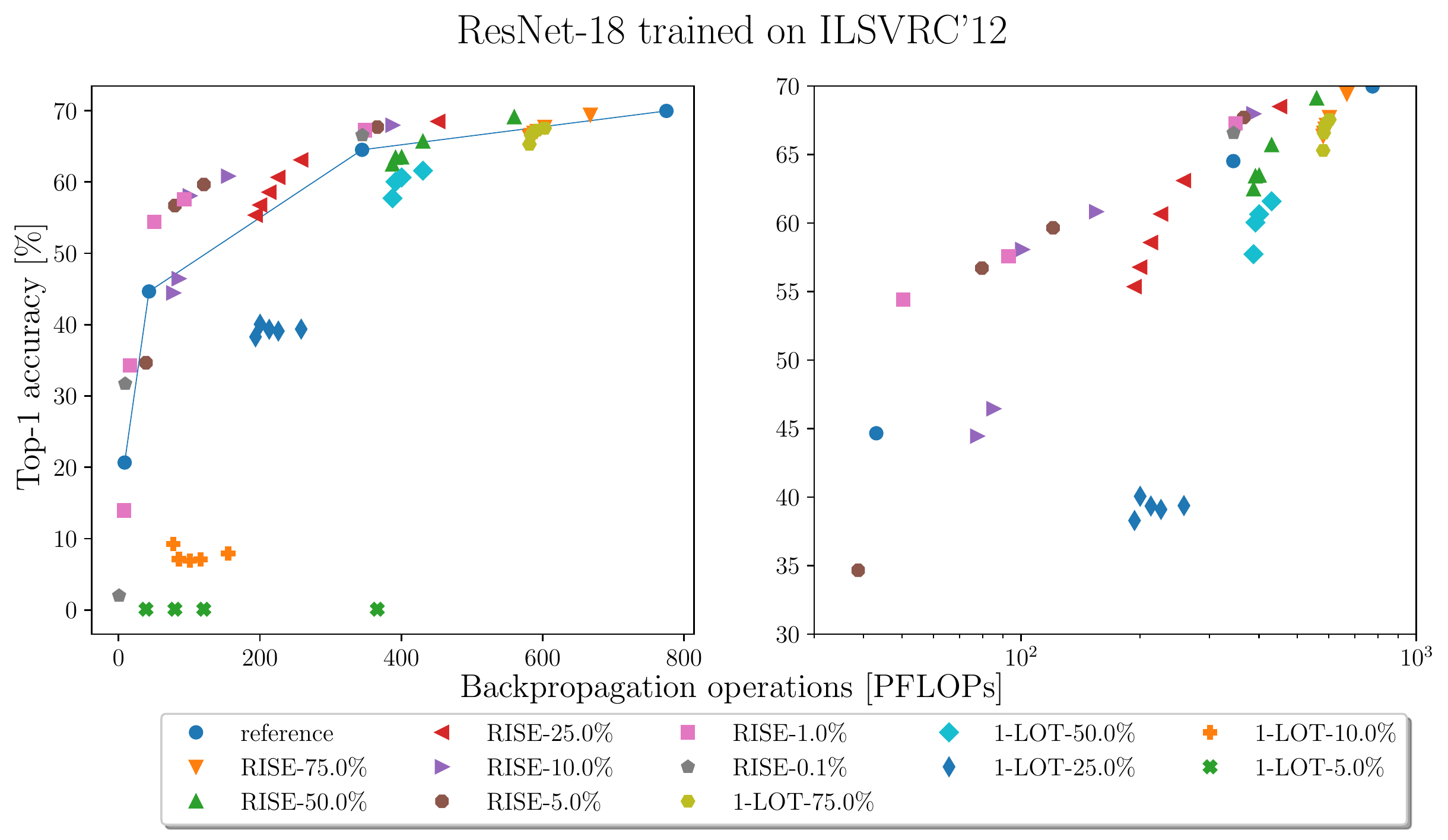}
\end{subfigure}
    
\caption{Training results for ResNet-32 trained on CIFAR-10 (top), MobileNet-v3 small on CIFAR-10 (center) and ResNet-18 on ILSVRC'12 (bottom).}
\label{fig:results}
\end{figure}

\textbf{The rise of the lottery heroes.} Fig.~\ref{fig:intuition} portraits the learning optimization constraint when pruning at initialization. When sampling the tickets and then rewinding, the model itself does not preserve the same initialization $W^k$, but it will be re-initialized a projection $W_{LOT}^k$, and its optimization is enforced in the subspace $\mathcal{W}$ (light blue). Despite such an approach does not introduce big problems in high $R$ regimes,\footnote{because the introduced perturbation $\Delta W^k$ in the initialization is small or, when $I$ grows, such perturbation is beneficially polarized towards removing parameters valuing approximately zero at the end of the training.} in low $R$ regimes the optimization problem is harder: the loss landscape becomes locally flat (Fig.~\ref{fig:hessianspecanalysis}) and the optimization problem can not be easily solved. However, we can ``lock'' the non-winning parameters and let the potential winners to rise and to evolve towards their final value, constraining the optimization problem for the values determined by $\mathcal{M}$ and freezing the others (light orange). Towards this end, we can modify the update rule in \eqref{eq:updatelottery} to
\begin{equation}
    \label{eq:updaterise}
    w_{i}^{t+1} = \left\{
    \begin{array}{ll}
    w_{i}^{t} - u_{i}^{t} & if~w_i \in \mathcal{W}\\
    w_{i}^{k}   & if~w_i \in \overline{\mathcal{W}}.
    \end{array}
    \right .
\end{equation}
Using this approach, we will no longer incur in the same obstacles as in Sec.~\ref{sec:lottery}, as we will optimize starting from the exact same loss landscape (Alg.~\ref{alg:rise}).\\
\textbf{Experiments.} In order to validate our approach, we run the following experiments: i) ResNet-32 trained on CIFAR-10 with same setup as described in Sec.~\ref{sec:lottery}; ii)MobileNet-v3 small in CIFAR-10 with training for 100 epochs with 5 epochs linear warm-up followed by cosine annealing (from learning rate 0.35), optimized with SGD with momentum 0.9 weight decay 6e-5 and batch size 128, learning rate tuning as in~\cite{mobnetcifar10}; iii) ResNet-18 on ILSVRC'12 with training for 90 epochs with initial learning rate 0.1 and decayed by a factor 0.1 at milestones 30 and 60, optimized with SGD with momentum 0.9 batch size 1024 and weight decay $5\cdot 10^{-5}$, same setup as in~\cite{resnetinet}. All the results are reported in Fig.~\ref{fig:results}. On the left the full results are displayed, on the right a zooming on the mostly dense regions is proposed, in log-scale. The continuous blue line is the reference training with the full model. Back-propagation operations are evaluated on the training complexity for one complete training. Every point in every graph represents a complete full training: the final performance achieved is reported. The multiple points with same color/shape refer to different $k$ value (refer to Alg.~\ref{alg:warmup} - for RISE line~\ref{line:calling_method} calls RISE-R): as $k$ increases, the back-propagation operations increase, as more training on the full model is required.\\ 
Unsurprisingly, we observe low performance for 1-LOT with low $R$, and despite different values of rewind, for low $R$ values the performance is heavily sub-optimal (like for $R=25\%$ in ResNet-32/CIFAR-10). On the contrary, even with extremely low $R$ regimes, we observe a progressive increment in the performance as $k$ increases. Notably, in the accuracy-backpropagation complexity plane, a Pareto-like curve is drawn by RISE: what emerges is that not the rewound epoch $k$, nor $R$ are really the metrics to determine the final performance of the model, but the training complexity deployed itself. Indeed, for low training complexity RISE achieves similar performance regardless of $R$ or $k$, under similar back-propagation complexity.
\section{Forecasting the rise of the lottery heroes?}
\label{sec:whatsnext}
In this work we have observed that traditional lottery ticket approaches are likely to fail in extreme scenarios when just a small subset of parameters is trained. However, locking the ``non-winning'' parameters and allowing the winners to evolve in the original loss landscape is a winning strategy. With such an approach it is possible to target a desired training performance training just a minimal portion of the entire model. In particular, the governing metrics in extreme regimes is the deployed training complexity. The results presented in this work, validated on standard architectures (ResNet), on already compact architectures trained with complex policies (MobileNet-v3) and on state-of-the-art datasets (ILSVRC'12) open the research towards the possibility of effectively deploying heavy computational saving at training time, as just a few directions are needed to train the model: the directions where the lottery heroes rise. Next work includes the identification of these directions at training time, as this work showed these exist and are algorithmically accessible. 

\bibliographystyle{IEEEbib}
\bibliography{main.bib}

\end{document}